\documentclass[journal,comsoc]{IEEEtran}
\usepackage[T1]{fontenc}% optional T1 font encoding

\usepackage{caption}
\usepackage{color}
\usepackage{ragged2e}
\usepackage{float}
\usepackage{array}
\usepackage{multirow}
\usepackage{url}
\usepackage{framed,multirow}
\usepackage{booktabs}

\ifCLASSOPTIONcompsoc
\usepackage[nocompress]{cite}
\else
\usepackage{cite}
\fi
\ifCLASSINFOpdf
   \usepackage[pdftex]{graphicx}
\else
\fi
\usepackage{amsmath}
\usepackage[linesnumbered]{algorithm2e}
\usepackage{color}
\ifCLASSOPTIONcompsoc
\usepackage[tight,normalsize,sf,SF]{subfigure}
\else
\usepackage[tight,footnotesize]{subfigure}
\fi
\begin{document}
%
% paper title
% Titles are generally capitalized except for words such as a, an, and, as,
% at, but, by, for, in, nor, of, on, or, the, to and up, which are usually
% not capitalized unless they are the first or last word of the title.
% Linebreaks \\ can be used within to get better formatting as desired.
% Do not put math or special symbols in the title.
\title{Multi-Agent Path Planning based on MPC and DDPG}

\author{Junxiao~Xue,
	Xiangyan~Kong,
	Bowei Dong,
	and~Mingliang~Xu% <-this % stops a space
	\thanks{J. Xue and X. Kong are with School of Software, Zhengzhou University.}% <-this % stops a space
	\thanks{B. Dong and M. Xu are with School of Information Engineering, Zhengzhou University.}}

% note the % following the last \IEEEmembership and also \thanks - 
% these prevent an unwanted space from occurring between the last author name
% and the end of the author line. i.e., if you had this:
% 
% \author{....lastname \thanks{...} \thanks{...} }
%                     ^------------^------------^----Do not want these spaces!
%
% a space would be appended to the last name and could cause every name on that
% line to be shifted left slightly. This is one of those "LaTeX things". For
% instance, "\textbf{A} \textbf{B}" will typeset as "A B" not "AB". To get
% "AB" then you have to do: "\textbf{A}\textbf{B}"
% \thanks is no different in this regard, so shield the last } of each \thanks
% that ends a line with a % and do not let a space in before the next \thanks.
% Spaces after \IEEEmembership other than the last one are OK (and needed) as
% you are supposed to have spaces between the names. For what it is worth,
% this is a minor point as most people would not even notice if the said evil
% space somehow managed to creep in.

% The paper headers
\markboth{Journal of \LaTeX\ Class Files,~Vol.~14, No.~8, August~2015}%
{Shell \MakeLowercase{\textit{et al.}}: Bare Demo of IEEEtran.cls for IEEE Communications Society Journals}
% The only time the second header will appear is for the odd numbered pages
% after the title page when using the twoside option.
% 
% *** Note that you probably will NOT want to include the author's ***
% *** name in the headers of peer review papers.                   ***
% You can use \ifCLASSOPTIONpeerreview for conditional compilation here if
% you desire.

% If you want to put a publisher's ID mark on the page you can do it like
% this:
%\IEEEpubid{0000--0000/00\$00.00~\copyright~2015 IEEE}
% Remember, if you use this you must call \IEEEpubidadjcol in the second
% column for its text to clear the IEEEpubid mark.

% use for special paper notices
%\IEEEspecialpapernotice{(Invited Paper)}

% make the title area
\maketitle

% As a general rule, do not put math, special symbols or citations
% in the abstract or keywords.
\begin{abstract}
The problem of mixed static and dynamic obstacle avoidance is essential for path planning in highly dynamic environment. However, the paths formed by grid edges can be longer than the true shortest paths in the terrain since their headings are artificially constrained. Existing methods can hardly deal with dynamic obstacles. To address this problem, we propose a new algorithm combining Model Predictive Control (MPC) with Deep Deterministic Policy Gradient (DDPG). Firstly, we apply the MPC algorithm to predict the trajectory of dynamic obstacles. Secondly, the DDPG with continuous action space is designed to provide learning and autonomous decision-making capability for robots. Finally, we introduce the idea of the Artificial Potential Field to set the reward function to improve convergence speed and accuracy. We employ Unity 3D to perform simulation experiments in highly uncertain environment such as aircraft carrier decks and squares. The results show that our method has made great improvement on accuracy by 7\%-30\% compared with the other methods, and on the length of the path and turning angle by reducing 100 units and 400-450 degrees compared with DQN (Deep Q Network), respectively.
\end{abstract}

% Note that keywords are not normally used for peerreview papers.

% For peer review papers, you can put extra information on the cover
% page as needed:
% \ifCLASSOPTIONpeerreview
% \begin{center} \bfseries EDICS Category: 3-BBND \end{center}
% \fi
%
% For peerreview papers, this IEEEtran command inserts a page break and
% creates the second title. It will be ignored for other modes.
\IEEEpeerreviewmaketitle

\section{Introduction}
% The very first letter is a 2 line initial drop letter followed
% by the rest of the first word in caps.
% 
% form to use if the first word consists of a single letter:
% \IEEEPARstart{A}{demo} file is ....
% 
% form to use if you need the single drop letter followed by
% normal text (unknown if ever used by the IEEE):
% \IEEEPARstart{A}{}demo file is ....
% 
% Some journals put the first two words in caps:
% \IEEEPARstart{T}{his demo} file is ....
% 
% Here we have the typical use of a "T" for an initial drop letter
% and "HIS" in caps to complete the first word.

\IEEEPARstart{I}{n} many dangerous and complex environments, path planning and obstacle avoidance still rely on manual decision-making. This method consumes a lot of human and material resources, and has the characteristics of low efficiency in path planning. Besides, in such environments like  path planning of carrier aircraft on carrier deck and path planning for post-disaster rescue, once humans make a mistake, it is likely to bring immeasurable losses and injuries. Hence, this requires agents to have the abilities of autonomous path planning and obstacle avoidance in complex environments. 

The core idea of path planning is to explore an optimal or sub-optimal barrier-free path according to certain evaluation criteria in obstacle environments \cite{1, 2}. Early path planning is mostly graph-based path planning \cite{3, 4, 5, 6}: such as V-graph, C-space \cite{3}, etc. However, graph-based methods cannot deal with the problem of dynamic obstacles. The Artificial Potential Field \cite{7}, Rapid-exploration Random Tree \cite{8} and D* \cite{9} can be used in dynamic environments. However, the Artificial Potential Field has the problem of local minimum. The Rapid-exploration Random Tree can’t deal with the continuous highly dynamic environments and the planned path is usually sub-optimal and not smooth. The D* algorithm is effective in finding a path in a dynamic environment. However, when moving to the target, it only checks the changes of the next node or neighboring nodes on the shortest path. So D* can’t deal with the changes of these nodes far away from the agent.

Reinforcement Learning (RL) comes out as a new research tendency that can grant the agent sufficient intelligence to make local decisions and accomplish necessary tasks. In \cite{10} and \cite{11}, the authors presented a Q-learning algorithm \cite{14} to solve the autonomous navigation problem of UAVs, afterwards, the Q-learning was also employed to establish paths while avoiding static or dynamic obstacles in \cite{12} and \cite{13}. Q-learning had good performance in these environments. However, in complex and highly dynamic environments, Q-learning hardly converges due to the curse of the dimensionality.

In recent years, artificial intelligence technology has been rapidly developed and applied. DRL \cite{17} was obtained by combining the advantages of Deep Learning (DL) \cite{15} and Reinforcement Learning (RL) \cite{16}, which provides a solution to the perceptual decision-making problem of complex systems. Liu et al. \cite{18} utilized the Deep Q Network (DQN) \cite{19} algorithm combined with greedy strategy to support agent trajectory planning and optimization in a large-scale complex environment. By using the value-based Dueling Double DQN (DDQN) \cite{20}, Zeng \cite{21} introduced a high-precision navigation technique for UAV, which has improved the efficiency of target tracking. However, these authors used discrete actions (i.e. the environment is modeled as a grid world with limited agent action space, degree of freedom), which may reduce the efficiency while dealing with real-world environment, where the agent operated according to a continuous action space.

To address these problems, in this paper, a new algorithm combining MPC \cite{24} (Model Predictive Control) with DDPG \cite{23} based on policy gradient \cite{22} is proposed. The main contributions of this research are as follows: 

\begin{itemize}
	\item A DDPG-based obstacle avoidance method is proposed, which enables agent to keep trial and error during the training process to learn a strategy to reach the destination without collision. In order to improve the efficiency of learning, the idea of the Artificial Potential Field (the obstacles and the target impose repulsion and attraction on agent respectively) is used to design the reward function.
	\item Considering the motion constraints of the agent in reality, the agent is modeled according to the Kinematic model. Then the MPC algorithm was used to predict the trajectory u(t) of the agent.
	\item The experimental results showed that our algorithm has made great improvement on accuracy by 7\%-30\% compared with the other algorithms, and on the length of the path and turning angle by reducing 100 units and 400-450 degrees compared with DQN (Deep Q Network), respectively.
\end{itemize}

The structure of this paper is as follows: Section 2, the path planning algorithm based on DDPG and MPC is introduced. Including Knowledge of background, Algorithm update, Reward function, State space and Action space. The simulation is presented in Section 3. We use unity3D to simulate the aircraft carrier deck and square. Section 4 gives the Experimental results that compared the model with DDPG, DQN, and A2C on the accuracy, length of path, reward and smoothness of the path. The summary is given in Section 5.

\section{ Path planning algorithm based on DDPG and MPC }
In this section, we present a new algorithm based on DDPG and MPC. Firstly, we introduce the knowledge of background including MPC, DRL and DDPG. Then we analyze algorithmic principle according to the algorithmic framework that is divided into three parts: trajectory prediction, action selection and update. Finally, the state space, action space and reward function under this framework are defined and explained.

\subsection{Knowledge of background}

\subsubsection{MPC}

Model predictive control (Model Predictive Control, MPC) is a type of control algorithm developed in the 1970s. The algorithm adopts control strategies such as multi-step prediction, rolling optimization and feedback correction. Its purpose is to minimize the cost function to make the prediction more accurate.
More precisely, in discrete time, the method of the model predictive control can be formulated as follows: 
$$ \begin{cases}
	\ x(t+1)=f(x(t),u(t),w(t)) \\
	\	y(t)=g(x(t),u(t)) \\
\end{cases}
\eqno{(1)} $$

Where $ u(t) $ is control matrix, $ w(t) $ is control error matrix caused by noise or disturbance, $ x(t) $ is the initial state, $ {y(t)}_{t}^{t+N} $ are the information of prediction calculated by the predictive model.

To make each prediction module and the predicted result closer to the true value, we need to perform rolling optimization on the model. The cost/objective function is defined as $J(\{x_{k}\}_{t}^{t+N},\{y_{k}\}_{t}^{t+N},\{u_{k}\}_{t}^{t+N})$. The optimal control input is generated by minimizing the cost function, that is:

$$ \{u_{k}^{*}\}_{t}^{t+N}=arg \mathop{min}_{\{u_{k}\}_{t}^{t+N}} J(\{x_{k}\}_{t}^{t+N},\{y_{k}\}_{t}^{t+N},\{u_{k}\}_{t}^{t+N})  \eqno{(2)}$$

The cost/objective function involve the future state trajectory $ \{x_{k}\}_{t}^{t+N} $, output trajectory $ \{y_{k}\}_{t}^{t+N} $ and control effort $ \{u_{k}\}_{t}^{t+N} $.

\begin{figure}[h]
	\centering
	\includegraphics[scale=0.4]{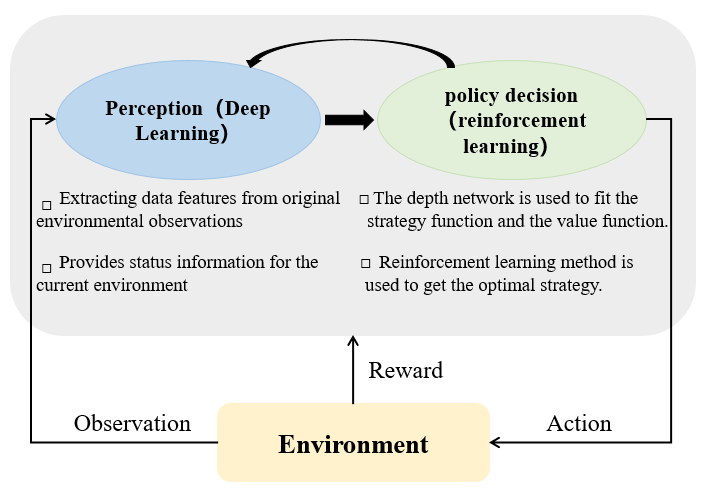}
	\caption{The framework of the DRL}
	\label{fig:Fig1}
\end{figure}

\subsubsection{Deep Reinforcement Learning}

\begin{figure*}[htp]
	\centering
	\includegraphics[scale=0.5]{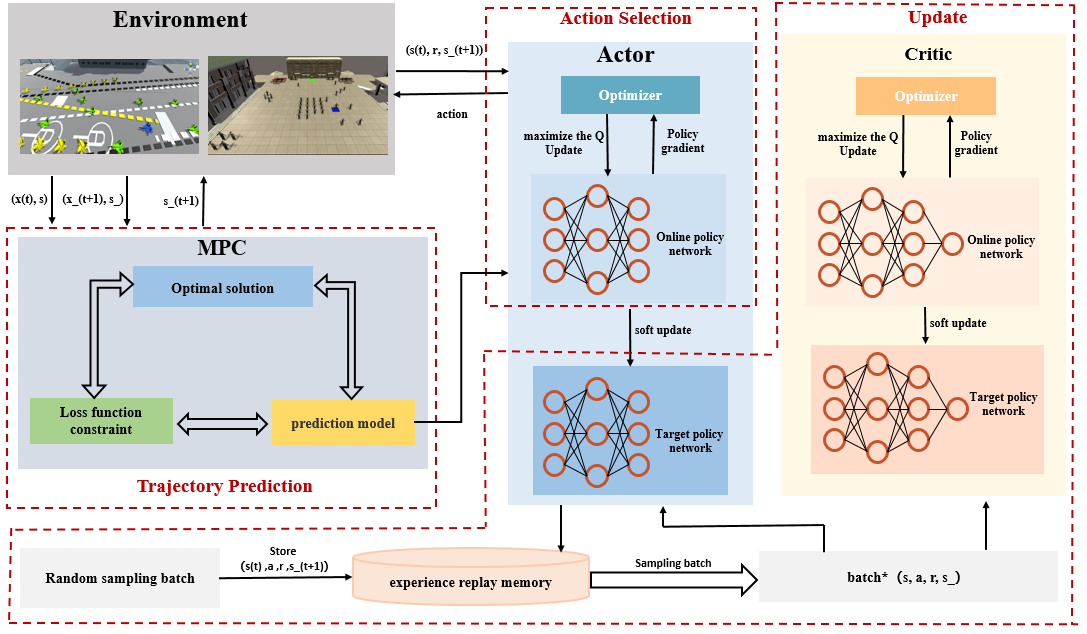}
	\caption{framework of our algorithm}
	\label{fig:Fig1}
\end{figure*}

DRL\cite{17} combines the perceptual ability of DL \cite{15} and the decision-making ability of RL\cite{16}, which can solve the high-dimensional decision-making problems. It is currently an extremely popular research direction in the field of machine learning. As shown in Fig.1, the framework of the DRL, the agent interacts with the environment to obtain a high-dimensional observation information, and the deep learning evaluates the value function of each action based on the expected return and completes the state-to-action mapping through a certain strategy. The environment reacts to this action and gets the next observation information.

\subsubsection{Deep Deterministic Policy Gradient}

DDPG is an extension of DQN based on the framework of Actor-Critic , which uses the experience replay sample collection method. It solves this problem that DQN can’t deal with continuous actions and Actor-Critic algorithm is difficult to converge. It adopts the four-network structure combining strategy gradient and value function. The strategy gradient is used to select a certain action from the continuous action space, and the value function is used to evaluate the selected action. 

DDPG adopts the delayed update method, however, unlike the hard update of DQN, it uses the method of soft update to update the target network. More precisely, the online policy Critic networks are updated by minimizing the mean square error. The online policy Critic networks are updated by maximizing the Q value. At last, the current networks parameter are updated by soft update strategies.

\subsection{Algorithmic Framework}

\subsubsection{Trajectory prediction}

Trajectory prediction refers to predict the short-term future coordinates according to history coordinates. In our algorithm, the MPC (Model Predictive Control) is used to predict the future coordinates, which has the advantages of good control effect and strong robustness.

As shown in Fig.1 the framework of our algorithm, the prediction model of MPC takes the current state s and the historical trajectory $ \{x_{k}\}_{t}^{t+N} $ of the dynamic obstacle as input and the output $ \{y_{k}\}_{t}^{t+N+1} $ are calculated by prediction model of MPC. In every time, the output $ y_{k} $ by prediction model and historical trajectory $ x_{k} $ are used to optimal the algorithm. The final predicted trajectory vector $ \{y_{k}\}^{t+N+1} $ and the current state of the environment vector $ s $ are concatenated as the final output.

\subsubsection{Action selection}

Action selection is the decision-making essence of reinforcement learning. As shown in Fig.1 The framework of our algorithm, the DDPG takes the predicted environmental state $ s(t) $ as input, and the output—action is calculated by online policy network of Actor. After the agent performs the action, the reward is given by our reward function. Afterwards, the current predicted environment, the reward, the action and the next predicted environment state  $(s(t),a_{t},r_{t},s(t+1)) $ are stored in the experience replay memory.  

\subsubsection{Update}

DDPG is an extension of DQN \cite{18} based on the AC \cite{24} framework. It retains the experience replay memory of DQN and dual network structure. However, unlike the hard update of DQN, it uses the soft update method to update the target network.

When updating the Actor and Critic networks, a mini-batch of N transitions are sampled from the experience replay memory.  Then the loss function $ L(\theta^{Q}) $ of Critic networks is calculated by the current target Q value $ y_{i} $. Meanwhile, the Actor network is updated by the method of policy gradient.
$$ y_{t}= \begin{cases}
	\ R_{t} & is\_end_{t} \; is \; true \\
	\ R_{t}+\gamma Q'(\phi(S_{t}'),\pi_{\theta '}(\phi(S_{t}'),\omega') & is\_end_{t} \; is \; false \\
\end{cases}  \eqno{(3)}$$

$$  L(\theta^{Q})=\frac{1}{N}\sum_{t}^N(Y_{i}-Q(s_{i},a_{i}|\theta^{Q}))^{2} \eqno{(4)}$$

$$ \nabla_{\theta^{\mu}}J\approx N^{-1}\sum_{t}^N\nabla_{a_{t}}Q(s_{t},a_{t}|\theta^{Q})\nabla_{\theta^{\mu}}\mu(s_{t}|\theta^{\mu}) \eqno{(5)}$$

At last, the two target network parameters $ \theta^{\mu'} $ and $ \theta^{Q'} $ are updated by soft update strategies:
$$  \begin{cases}
	\ \theta^{Q'}= \tau\theta^{Q}+(1-\tau)\theta^{Q'} \\
	\	\theta^{\mu'}= \tau\theta^{\mu'}+(1-\tau)\theta^{\mu'} \\
\end{cases}
\eqno{(6)}$$

Where $ \tau $ is a configurable constant coefficient, used to the regulate the soft update factor.

\subsection{State space}

The state space of agent represents valuable information agent can attain before decision-making, which is used to help agent assesses the situation. For the agent to better understand the changing environment, we divided the state into the current environmental state and the predicted environmental state. The status value can be illustrated as $ s(t):\{s,u(t)\} $.

\subsubsection{The current environmental state}
$$ s:\quad \{(d_{x}^{p},d_{y}^{p}),(d_{x}^{o1},d_{y}^{o1})\ldots (d_{x}^{om},d_{y}^{om})\}  \eqno{(7)}$$

Where $ (d_{x}^{p},d_{y}^{p}) $ represents the distance between the coordinates of the agent point and the target, and the $ (d_{x}^{o1},d_{y}^{o1})-(d_{x}^{om},d_{y}^{om}) $ represent the distance between the coordinates of the agent and the static obstacles.

\subsubsection{The predicted environmental state}
$$ u(t):\quad \{(d_{x}^{o'1},d_{y}^{o'1}),(d_{x}^{o'2},d_{y}^{o'2})\ldots (d_{x}^{o'n},d_{y}^{o'n})\} \eqno{(8)}$$

Where $(d_{x}^{o'1},d_{y}^{o'1})-(d_{x}^{o'n},d_{y}^{o'n}) $ represent the distance between the predicted coordinates of the dynamic obstacles and the coordinates of the agent.	

\subsection{Agent action space}

The action space represents the set of actions that the agent can perform. We set the action space as:
$$ A:(x,y) \quad x,y\in(-1,1) \eqno{(9)}$$

The distances that the agent moves in the x and y directions are expressed as: $ X=x*40 $ $ Y=y*40 $, respectively.

\begin{figure*}[htbp]
	\centering
	\begin{minipage}[t]{0.48\textwidth}
		\centering
		\includegraphics[width=6.5cm]{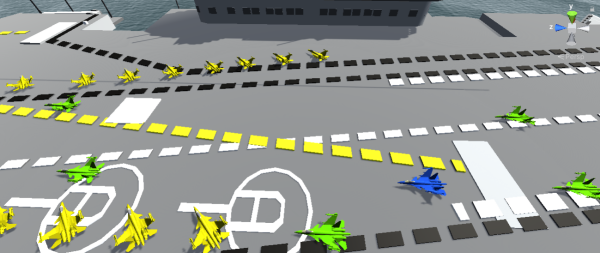}
		\caption{Scene 1}
	\end{minipage}
	\begin{minipage}[t]{0.48\textwidth}
		\centering
		\includegraphics[width=6.5cm]{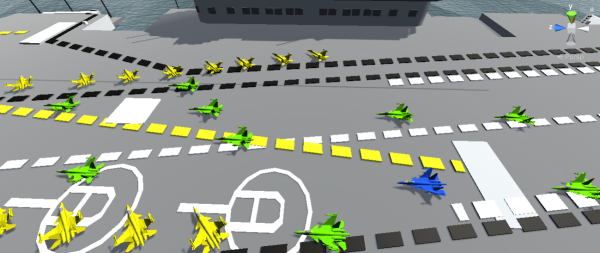}
		\caption{Scene 2}
	\end{minipage}
	\begin{minipage}[t]{0.48\textwidth}
		\centering
		\includegraphics[width=6.5cm]{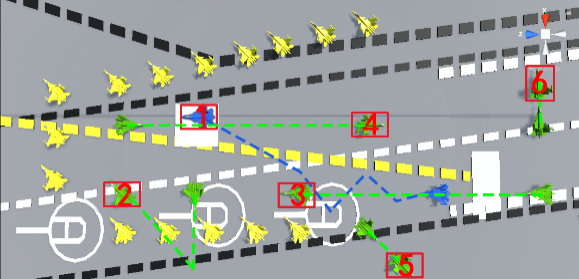}
		\caption{trajectory of Scene 1}
	\end{minipage}
	\begin{minipage}[t]{0.48\textwidth}
		\centering
		\includegraphics[width=6.5cm]{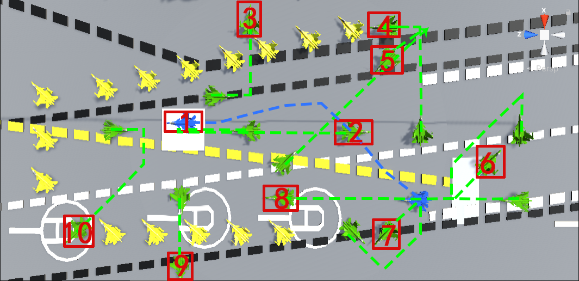}
		\caption{trajectory of Scene 2}
	\end{minipage}
	\begin{minipage}[t]{0.48\textwidth}
		\centering
		\includegraphics[width=6.5cm]{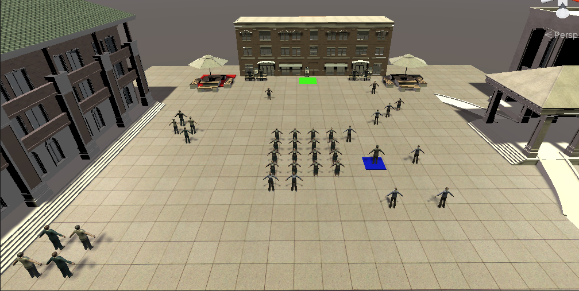}
		\caption{Scene 3 simulation diagram}
	\end{minipage}
	\begin{minipage}[t]{0.48\textwidth}
		\centering
		\includegraphics[width=6.5cm]{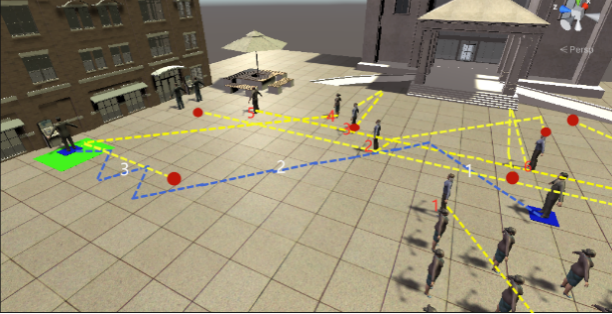}
		\caption{Scene 3 simulation diagram}
	\end{minipage}
\end{figure*}

\subsection{Reward function}

\begin{figure*}[ht]
	\centering
	\subfigure[DQN] {
		\label{fig:a}
		\includegraphics[scale=0.4]{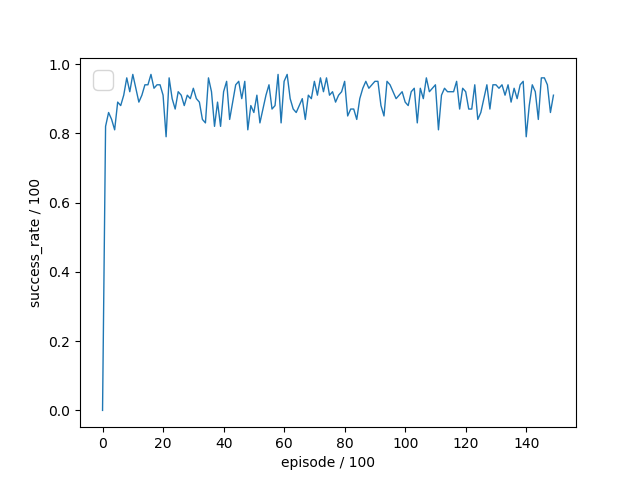}
	}
	\hspace{0.3in}
	\subfigure[A2C] {
		\label{fig:b}
		\includegraphics[scale=0.4]{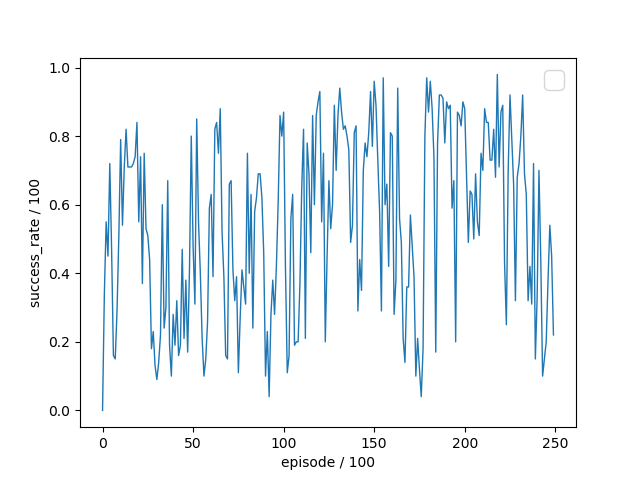}
	}
	\hspace{0.3in}
	\subfigure[DDPG] {
		\label{fig:c}
		\includegraphics[scale=0.4]{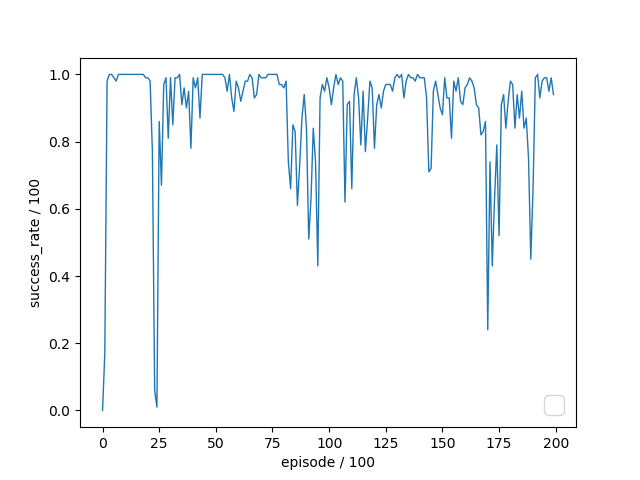}
	}
	\subfigure[our algorithm] {
		\label{fig:d}
		\includegraphics[scale=0.4]{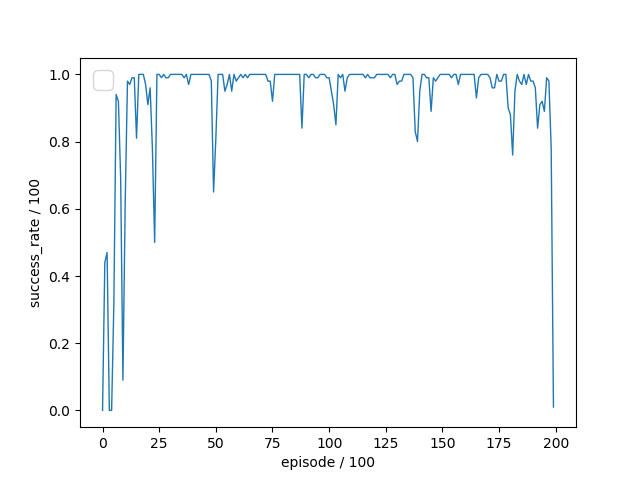}
	}
	
	\caption{ The accurate rate of scene 1 }
	\label{fig}
\end{figure*}

\begin{figure*}[ht]
	\centering
	\subfigure[DQN] {
		\label{fig:a}
		\includegraphics[scale=0.4]{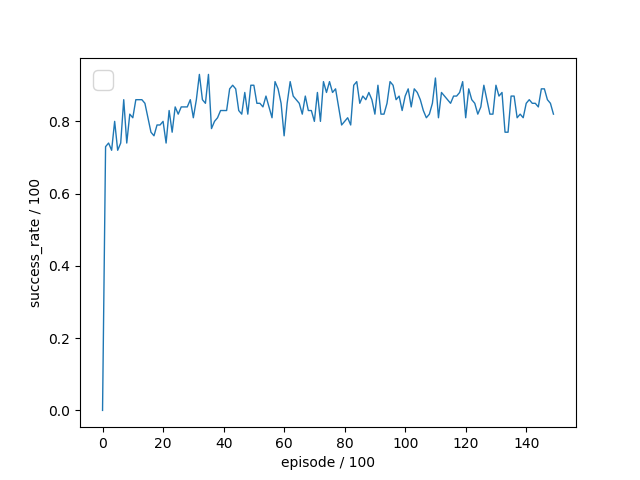}
	}
	\hspace{0.3in}
	\subfigure[A2C] {
		\label{fig:b}
		\includegraphics[scale=0.4]{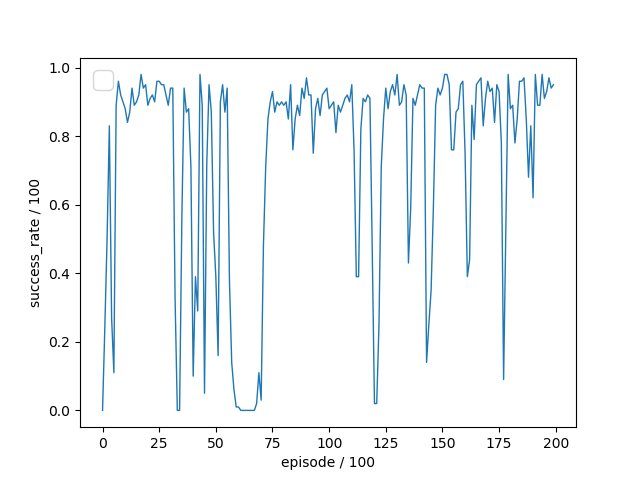}
	}
	\hspace{0.3in}
	\subfigure[DDPG] {
		\label{fig:c}
		\includegraphics[scale=0.4]{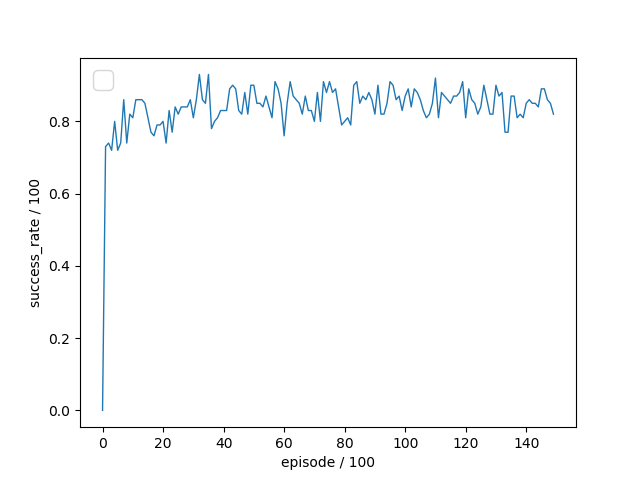}
	}
	\subfigure[our algorithm] {
		\label{fig:d}
		\includegraphics[scale=0.4]{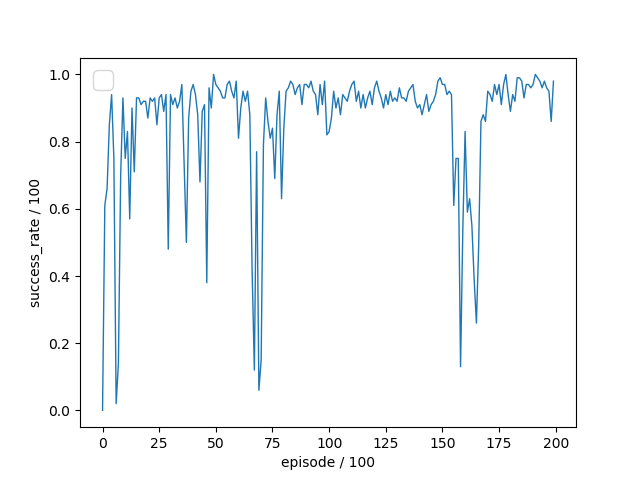}
	}
	
	\caption{ The accurate rate of scene 2 }
	\label{fig}
\end{figure*}

The reward in DRL, which is the feedback signal available for the agent’s learning, is used to evaluate the action taken by the agent. A simple approach is to set sparse reward, in which the agent can get a positive return only if the task is accomplished. However, this method is inefficient to collect useful experience data so as to help agent learning.  Accordingly, the convergent speed of network updating is slow and the agent may not learn optimal strategies. In this paper, we introduce the idea of the Artificial Potential Field (the obstacles and the target impose repulsion and attraction on agent respectively) into the design of reward function. Four types of excitation are considered in the reward. (1) The reward of target attraction. (2) The reward of obstacle repulsion. (3) Collision reward. (4) Reward for reaching the target point. More precisely, these rewards are described as follows:

\subsubsection{The reward of target attraction}

The magnitude of attraction is proportional to the distance between the agent and the target. As shown in Eq. (10) and (11), we simplified the attraction of the Artificial Potential Field.
$$ r_{1}= \begin{cases}
	\ L & D_{i,e}^{t}-D_{i,e}^{t+1}\geq L \\
	\ D_{i,e}^{t}-D_{i,e}^{t+1} & l<D_{i,e}^{t}-D_{i,e}^{t+1}<L \\
	\ l & D_{i,e}^{t}-D_{i,e}^{t+1}\leq l \\
\end{cases}  \eqno{(10)}$$

$$ r_{1}= \begin{cases}
	\ -l & D_{i,e}^{t}-D_{i,e}^{t+1}\geq -l \\
	\ D_{i,e}^{t}-D_{i,e}^{t+1} & -L<D_{i,e}^{t}-D_{i,e}^{t+1}<-l \\
	\ -L & D_{i,e}^{t}-D_{i,e}^{t+1}\leq -L \\
\end{cases}  \eqno{(11)}$$

The reward of target attraction $ r_{1} $ is usually proportional to the difference between $ D_{i,e}^{t} $ (the distance between the agent and the target at time t) and $ D_{i,e}^{t+1} $ (he distance between the agent and the target at time t+1). However, when the difference is too large or too small, $ L,l $ or $ -l,-L $ replace the difference as the reward. Among them, the formulas (10) and (11) represent agent is close or far away from the target respectively.

\subsubsection{ The reward of obstacle repulsion}

Similar to the reward of target attraction , we also simplified the repulsion value of the Artificial Potential Field. The formulas as shown in Eq. (12) and (13).
$$ r_{2}= \begin{cases}
	\ H & D_{i,ob}^{t}-D_{i,ob}^{t+1}\geq H \\
	\ D_{i,ob}^{t}-D_{i,ob}^{t+1} & h<D_{i,ob}^{t}-D_{i,ob}^{t+1}<H \\
	\ h & D_{i,ob}^{t}-D_{i,ob}^{t+1}\leq h \\
\end{cases}  \eqno{(12)}$$

$$ r_{2}= \begin{cases}
	\ -h & D_{i,ob}^{t}-D_{i,ob}^{t+1}\geq -h \\
	\ D_{i,ob}^{t}-D_{i,ob}^{t+1} & -H<D_{i,ob}^{t}-D_{i,ob}^{t+1}<-h \\
	\ -H & D_{i,ob}^{t}-D_{i,ob}^{t+1}\leq -H \\
\end{cases}  \eqno{(13)}$$

Where $ D_{i,ob}^{t} $ and $ D_{i,ob}^{t+1} $ represent the distance between the agent and the obstacle at time $ t $ and $ t+1 $ respectively. $ H,h $ or $ -h,-H $ represent the upper and lower limits of the reward value respectively when agent is close or far away from obstacles.

\subsubsection{Collision reward}

As shown in Eq. (14), the reward is given by the environment when the agent collides with the obstacles.
$$ r_{3}=-50   \eqno{(14)}$$

\subsubsection{Reward for reaching the target point}

As shown in Eq. (15), the reward is given by the environment when the agent reach to the target .
$$ r_{4}=200   \eqno{(15)}$$

At last, the total reward value can be defined in Eq. (16).
$$ R=r_{1}\ast\lambda_{1}+r_{2}\ast\lambda_{2}+r_{3}\ast\lambda_{3}+r_{4}\ast\lambda_{4}  \eqno{(16)}$$

Where $ \lambda_{1},\lambda_{2},\lambda_{3},\lambda_{4} $ represents the weight of the four kinds of reward values respectively.

\section{ Simulation }

\begin{figure*}[ht]
	\centering
	\subfigure[DQN] {
		\label{fig:a}
		\includegraphics[scale=0.4]{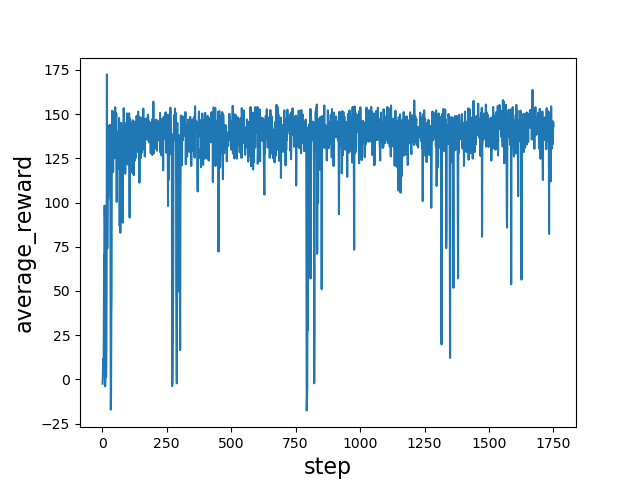}
	}
	\hspace{0.3in}
	\subfigure[A2C] {
		\label{fig:b}
		\includegraphics[scale=0.4]{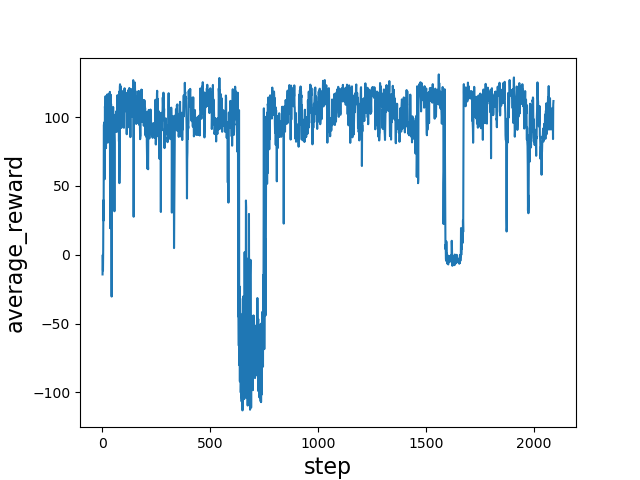}
	}
	\hspace{0.3in}
	\subfigure[DDPG] {
		\label{fig:c}
		\includegraphics[scale=0.4]{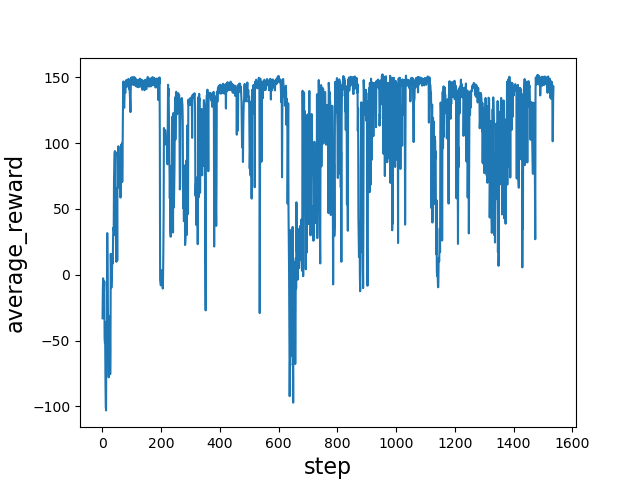}
	}
	\subfigure[our algorithm] {
		\label{fig:d}
		\includegraphics[scale=0.4]{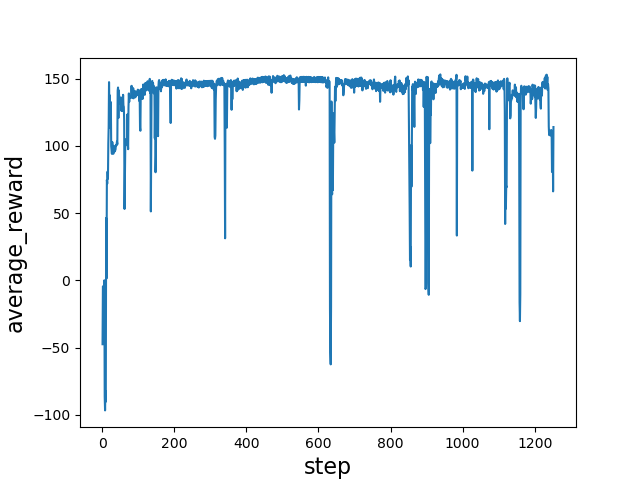}
	}
	
	\caption{ The average reward of scene 1 }
	\label{fig}
\end{figure*}

\begin{figure*}[ht]
	\centering
	\subfigure[DQN] {
		\label{fig:a}
		\includegraphics[scale=0.4]{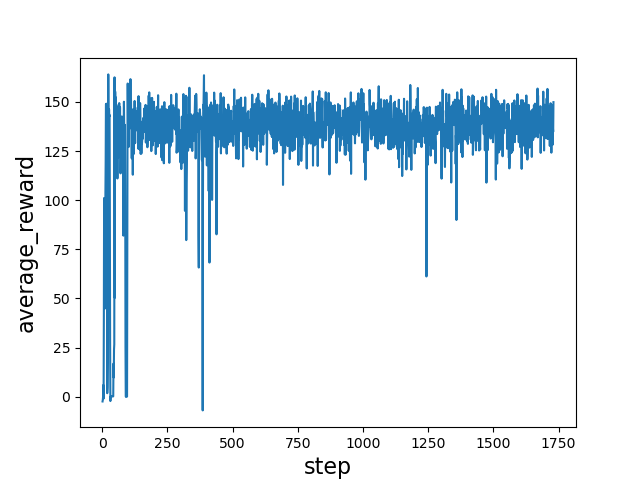}
	}
	\hspace{0.3in}
	\subfigure[A2C] {
		\label{fig:b}
		\includegraphics[scale=0.4]{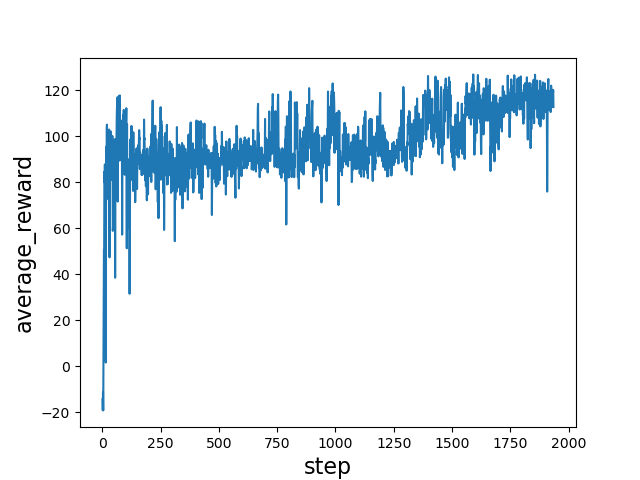}
	}
	\hspace{0.3in}
	\subfigure[DDPG] {
		\label{fig:c}
		\includegraphics[scale=0.4]{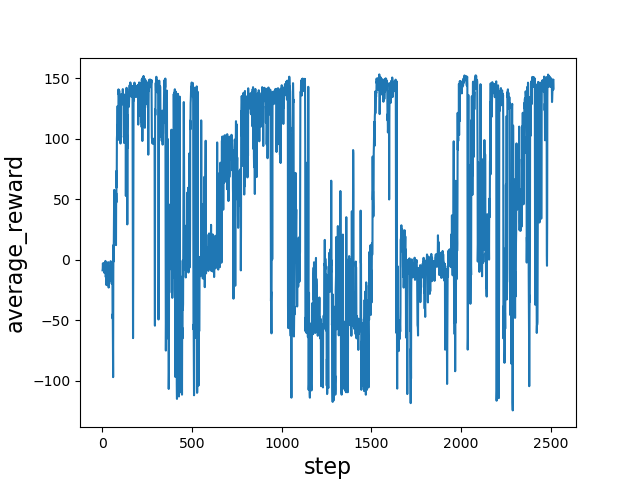}
	}
	\subfigure[our algorithm] {
		\label{fig:d}
		\includegraphics[scale=0.4]{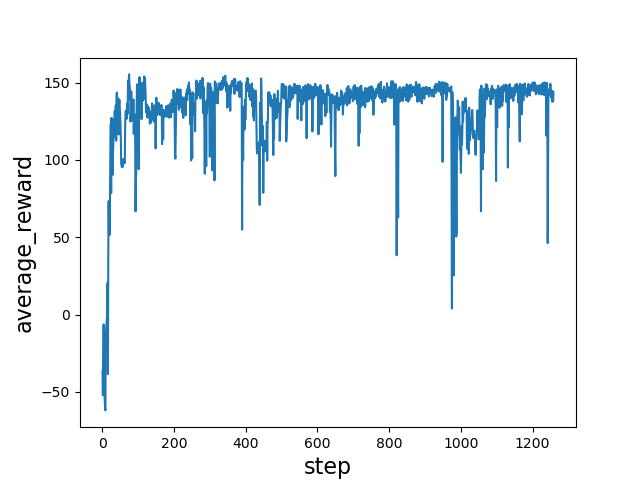}
	}
	
	\caption{ The average reward of scene 2 }
	\label{fig}
\end{figure*}

The path planning of the agent and dynamic obstacle avoidance are important problems in computer simulation research and they are widely used in crowd animation, computer game and deduction of scheme. Among them, the pedestrian square has the characteristics of large area, high density poor regularity in crowd movement and high mobility. While, the deck space of the aircraft carrier is narrow and tasks are numerous and the uninterrupted operation of carrier-based aircraft increases the uncertainty of space usage. It is a challenge to find a collision-free path from the starting point to the target in such highly uncertain environments. Therefore, in this section, we chose the aircraft carrier deck and square as the objects of simulation to verify the efficiency of our algorithm, and the Unity3D is used to model this simulation. Meantime, we sampled randomly an episode date to described the trajectories of the carrier-based aircraft and dynamic obstacles in every scenes.

\subsection{Aircraft carrier deck simulation}

As shown in Fig.2 and Fig.3, in the aircraft carrier deck, the blue carrier-based aircraft needs to avoid dynamic and static obstacles (the green and yellow carrier-based aircrafts represented dynamic and static obstacles, respectively) in real-time to find a collision-free path from the start point to the target (the position of the white board). In scene 1, we set 5 dynamic obstacles and 15 static obstacles. As shown in Fig.3, in order to verify the effective of our algorithm in the highly dynamic environment, 4 dynamic obstacles were added randomly to the scene 1.

To simulate the movement of dynamic obstacles the real scene. Three assumptions were made as follows:

Assumption 1: The movement of the dynamic obstacle have the purpose;

Assumption 2: Special locations are set as the target of dynamic obstacles in the scene (for example gas station, command room, etc.);

Assumption 3: The initial speeds are set to dynamic obstacles. Then when it moves to a special position, it will stay there for several time to perform operations, the time of operation is given randomly. After that, a new speed of dynamic obstacles is given randomly.

For two scenes, we described the trajectories of the carrier-based aircraft and dynamic obstacles in an episode. As shown in in Fig.4 and Fig.5, the trajectories of the carrier-based aircraft and dynamic obstacles were represented by the blue and green dashed lines, respectively. The final location of carrier-based aircraft and dynamic obstacles were marked with the red rectangle and numbers to identify them. In scene 1, when the carrier-based aircraft predicted that it would collide with the obstacle 3, an action——upward was taken to avoid the obstacle successfully in real-time. After that, the carrier-based aircraft predicted that the collision would be occurred with the obstacle 4 and then the direction was adjusted to avoid. In scene 2, when obstacles 2 and 5 appeared around the carrier-based aircraft, the carrier-based aircraft would move upward to avoid two dynamic obstacles, and ensured that the carrier-based aircraft moves closer to the target.

\subsection{Crowd square simulation}

As shown in Fig.6, in the pedestrian square, the people on the blue square was moving toward the target to withdraw money in   the position of green square board. In  this scene, we set 28 static obstacles and 13 dynamic obstacles. It was worth noting that 9 dynamic obstacles and a large number of static obstacles were set around the agent to improve the difficulty of obstacle avoidance.

The trajectories of pedestrians and agent were depicted on the square in an episode by the blue and yellow dashed lines, respectively in Fig. 5. It is obvious that the agent moved following trajectory 1 to avoid the static crowd and pedestrian 1. Then the agent predicted that if continue to move in the original direction, it would collide with pedestrian 2 and deviate from the target, so the agent adjusted its direction and moved along with trajectory 2 to avoid collision and move towards the target. When the agent arrived near the target, pedestrian 5 finished the task of money withdrawal and moved outward. After the agent predicted that it would collide with pedestrian 5, he adjusted constantly the direction of movement to avoid obstacles and reached the target without collision in finally.

\section{Results}

For avoiding repetition and verifying the efficiency of the model, in this section we just adopted the method of comparative experiment on the aircraft carrier deck scene, and systematically compared the model with DDPG, DQN, and A2C on the accuracy rate, length of path, reward and smoothness of the path. In particular, the smoothness of the path was expressed by the total turning angle of the aircraft reaching the target point without collision.

First of all, we compared the accuracy rate for the different algorithms in every scene. In scene 1, as shown in Fig.8, the accuracy rate of our algorithm was easiest to reach 100\% in four algorithms. Due to the difficulty of convergence of the A2C, the accuracy rate of A2C was the most unstable. Secondly, although the accuracy rate of DDPG could reach 100\%, it still fluctuated greatly. Although DQN had a good performance in accuracy rate, it still has disadvantages, which would be introduced in the later. To compare the results more clearly, we calculated the average accuracy rate and showed it in the table. As shown in Tables 1, 100 sets of accuracy rate were sampled  randomly to calculate the average, our algorithm made great improvement on accuracy rate by 8\%, 6\%, and 31\% compared with DQN, A2C and DDPG, respectively. In scene 2, with the complexity of the environment increases, the accuracy rate of the four algorithms had decreased, but our algorithm could still reach 90\%. Our algorithm still had the highest accuracy rate in four algorithms. In summary, in terms of the accuracy rate of the algorithm, although both DDPG and A2C can solve the problems in the continuous action space, DDPG has a better learning effect. Our algorithm combines the prediction module on the basis of DDPG, so that the algorithm can  achieve a higher accuracy rate in complex environments.

\begin{table}[h]   \caption{\label{tab:test}Scene1 Algorithm comparison table}  \centering  \begin{tabular}{ccc}    \toprule   
		&  Scene 1 & Scene 2  \\    \midrule  
		DQN &  90\% & 84\% \\
		A2C &  67\% & 64\% \\
		DDPG &  92\% & 82\% \\
		MDDPG & 98\% & 91\% \\
		\bottomrule   \end{tabular}  \end{table}

To verify the learning effect of the algorithm, we used our algorithm to compare the average reward with other algorithm. The average reward of the four algorithms in two scenes were presented in Fig.10 and Fig.11. The average reward indicated the efficiency of the learning. As shown in Fig. 9, we compared the average reward of the four algorithms in scene 1. Due to the difficulty of convergence of the A2C algorithm, the maximum average reward would only stabilize at about 100. However, DQN, DDPG, and our algorithm would reach 150. It was obvious that the average reward of DDPG was unstable. That showed that it was difficult for DDPG to deal with dynamic obstacles without prediction. Although the curve of the average reward is similar between DQN and our algorithm, our algorithm had higher stability and was more in line with the kinetic model in the real scene. It could be seen from the average reward that our algorithm had the best learning effect in four algorithms.

\begin{table}[h]   \caption{\label{tab:test}Scene 1 Algorithm comparison table}   \begin{tabular}{ccc}    \toprule   
		&  Length of path & Turning angle  \\    \midrule  
		DQN & 438.22 & 586.73 \\
		A2C & 352.60 & 142.03 \\
		DDPG & 328.97 & 139.96 \\
		MDDPG & 328.95 & 139.80 \\
		\bottomrule   \end{tabular}  \end{table}

\begin{table}[h]   \caption{\label{tab:test}Scene 2 Algorithm comparison table}   \begin{tabular}{ccc}    \toprule   
		&  Length of path & Turning angle  \\    \midrule  
		DQN &  465.45 & 522.62 \\
		A2C &  324.16 & 158.00 \\
		DDPG &  320.61 & 134.28 \\
		MDDPG & 315.68 & 128.41 \\
		\bottomrule   \end{tabular}  \end{table}

To take into account as much as possible the planning effect of the algorithm in the real scene, we compared the average length of the path and average turning angle of the algorithm. In a real scene, finding a smooth and optimal or sub-optimal path can not only reduce the consumption of resource, but also complete the task in the shortest time, increasing the real-time nature of the task. As shown in Table 2 and Table 3, which indicated the average length of the path and average turning angle of the four algorithms respectively in two different scenes. Although DQN had a good performance in accuracy rate, the average length of the path and average turning angle were longest in four algorithms, which would consume more time and resources in the real environment. Meanwhile, we compared two different scenes. Though the accuracy rate of the four algorithms had reduced after increasing the complexity of the environment. The average length of the path and average turning angle had small fluctuations.

In summary, our algorithm could effectively solve the difficult convergence of A2C and the difficulty of DDPG to deal with dynamic obstacles. Compared with the other three algorithms, our algorithm has obvious improvements in terms of average path length, path smoothness, accuracy and average reward. 

\section{Conclusion}

In this paper, a new algorithm of path planning based on DDPG and MPC is designed which combines the perception and decision-making capabilities of deep reinforcement learning and the predictive capabilities of the MPC, and can apply to the highly uncertain scene. The trajectory prediction of dynamic obstacles is achieved by the MPC, which greatly reduces the uncertainty of the environment, and effectively solves the problems faced by traditional algorithms in dynamic environments, such as slow convergence speed and poor generalization. The DDPG is suitable for solving the  problems of continuous state space. which is more in line with real scenes. In order to verify the efficiency of the algorithm, we chose Unity3D to simulate the complex aircraft carrier deck and square, and conducted a detailed analysis of the agent's trajectory. The final results show that our algorithm has made great improvement on accuracy by 7\%-30\% compared with the other algorithms, and on the length of the path and turning angle by reducing 100 units and 400-450 degrees compared with DQN (Deep Q Network), respectively.

\bibliographystyle{IEEEtran}

\bibliography{bare_jrnl_comsoc}

%\begin{IEEEbiography}[{\includegraphics[width=1in,height=1.25in,clip,keepaspectratio]{photo/XueJX.jpg}}]{Junxiao Xue}
%	is an associate professor in School of Software, Zhengzhou University, China. His research interests include virtual reality and crowd simulation. He received his Ph.D in 2009 from the School of Mathematical Sciences, Dalian University of Technology, China.
%\end{IEEEbiography}

%\begin{IEEEbiography}[{\includegraphics[width=1in,height=1.25in,clip,keepaspectratio]{photo/HuiYin.jpg}}]{Hui Yin}

%\end{IEEEbiography}

%\begin{IEEEbiography}[{\includegraphics[width=1in,height=1.25in,clip,keepaspectratio]{photo/XuMingliang.jpg}}]{Mingliang Xu}
%	is a full professor in the School of Information Engineering of Zhengzhou University, China, and currently is the director of CIISR (Center for Interdisciplinary Information Science Research) and the vice General Secretary of ACM SIGAI China. He received his Ph.D. degree in computer science and technology from the State Key Lab of CAD\&CG at Zhejiang University, Hangzhou, China. He previously worked at the department of information science of NSFC (National Natural Science Foundation of China), Mar.2015-Feb.2016. His current research interests include computer graphics, multimedia and artificial intelligence. He has authored more than 60 journal and conference papers in these areas, including ACM TOG, ACM TIST, IEEE TPAMI, IEEE TIP, IEEE TCYB, IEEE TCSVT, ACM SIGGRAPH (Asia), ACM MM, ICCV, etc.
%\end{IEEEbiography}
\end{document}